# Topic Sensitive Neural Headline Generation

**Lei Xu**,[*] **Ziyun Wang**,[*] **Ayana, Zhiyuan Liu**,[†] **Maosong Sun**[‡]
State Key Laboratory of Intelligent Technology and Systems
Tsinghua National Laboratory for Information Science and Technology
Department of Computer Science and Technology, Tsinghua University, Beijing, China

## Abstract

Neural models have recently been used in text summarization including headline generation. The model can be trained using a set of document-headline pairs. However, the model does not explicitly consider topical similarities and differences of documents. We suggest to categorizing documents into various topics so that documents within the same topic are similar in content and share similar summarization patterns. Taking advantage of topic information of documents, we propose topic sensitive neural headline generation model. Our model can generate more accurate summaries guided by document topics. We test our model on LCSTS dataset, and experiments show that our method outperforms other baselines on each topic and achieves the state-of-art performance.

## 1 Introduction

Text summarization, including headline generation, is an important task in natural language processing. It is typically challenging to capture the core information of a document and create an informative but brief summary of the document.

Most existing text summarization approaches can be divided into two categories: extractive and generative. Extractive summarization (Edmundson, 1969) simply selects a few sentences from the given document and reorder them into a compact summary. Due to the limitation of vocabulary and sentence structure, it is extremely difficult for extractive models to generate coherent and concise summaries. Generative models, on the other hand, aim at comprehending a document and generating the summary not necessarily having appeared in the original document.

Recent years have witnessed the development of sequence-to-sequence (seq2seq) neural models (Sutskever et al., 2014). These models typically learn distributed representations (Le and Mikolov, 2014) of an input sequence, and then generate an output sequence accordingly. The advantage of neural models is that these models learn a semantic mapping directly according to pairs of document-headline sequences without designing hand-crafted features.

Neural models have shown great superiority for text summarization (Rush et al., 2015) and headline generation, because these models can flexibly model document semantics from internal word sequences within the document. Nevertheless, much external information about documents may also play important roles for text summarization. For example, documents usually group into various topics, and the documents within a certain topic may exhibit specific summarization patterns. For example, a document about *economy* is usually summarized including causes and effects; and a document about *social events* or *accidents* is always summarized containing time, location and place of the event.

In this paper, we propose to incorporate topic information of documents into neural models for text summarization and propose topic-sensitive neural headline generation (TopicNHG). More specifically, we apply Latent Dirichlet Allocation (LDA) (Blei et al., 2003) to assign topic labels for documents, and introduce the topic labels in neural models to build unique encoders and decoders for each topic respectively. In this way, TopicNHG can effectively identify the corresponding crucial parts in a document guided by its topic information, and are expected to generate well-focused headlines.

[*]Indicates equal contribution.
[†]liuzy@tsinghua.edu.cn
[‡]sms@tsinghua.edu.cn

In this paper, we evaluate our model on a large-scale Chinese LCSTS dataset (Baotian Hu, 2015). Experiment results show that our model significantly outperforms other baseline systems. Moreover, it consistently performs the best on each individual topic, which proves the statistical significance and robustness of TopicNHG.

## 2 Neural Headline Generation

Neural headline generation aims at learning a model to map a short text into a headline. The model uses a gated recurrent unit (GRU) encoder to learn a feature vector from the input text, based on which a GRU decoder generates a concise headline.

### 2.1 Gated Recurrent Unit

GRU (Chung et al., 2014) is an extension of Recurrent neural networks (RNN) (Mikolov et al., 2010). GRU processes an input sequence $\mathbf{x} = \{x_1, x_2, \ldots, x_{|\mathbf{x}|}\}$ and generates a sequence of output states $\mathbf{h} = \{h_1, h_2, \ldots, h_{|\mathbf{x}|}\}$. The $t$-th unit is fed with previous output $h_{t-1}$ and current input $x_t$ and produces its output $h_t$. When calculating $h_t$, it uses update gate $z_t$ and reset gate $r_t$ to improve the performance on long sequences. The gates are computed as

$$z_t = \sigma(\mathbf{W}_z x_t + \mathbf{U}_z h_{t-1}), \quad (1)$$

$$r_t = \sigma(\mathbf{W}_r x_t + \mathbf{U}_r h_{t-1}), \quad (2)$$

where $\sigma(\cdot)$ is sigmoid function. Then it computes candidate output $\hat{h}_t$ and final output $h_t$ as

$$\hat{h}_t = \tanh(\mathbf{W}_h x_t + \mathbf{U}_h(r_t \cdot h_{t-1})), \quad (3)$$

$$h_t = (1 - z_t) \cdot h_{t-1} + z_t \cdot \hat{h}_t. \quad (4)$$

### 2.2 Encoder-Decoder

NHG includes an encoder to encode input text $\mathbf{x}$ into a feature vector $v$ and a decoder to generate headline $\mathbf{y}$ based on $v$. The attention mechanism can improve the performance (Bahdanau et al., 2014). It assigns different feature vectors $v_t$ for different steps of the decoder. Fig. 1 shows the framework of NHG. The model generates the output sequence following a Markov process, which means

$$p(\mathbf{y}|\mathbf{x}, \theta) = \prod_{t=1}^{n} p(y_t|\mathbf{x}, \mathbf{y}_{1:i-1}, \theta), \quad (5)$$

where $\mathbf{y}_{1:i}$ means $\{y_1, y_2, \ldots, y_i\}$, and $\theta$ is the model parameter.

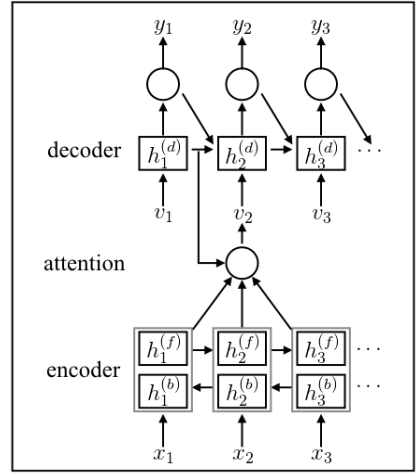

**Figure 1:** NHG with Attention Mechanism.

**Encoder** The encoder is a bidirectional GRU which process $\mathbf{x} = \{x_1, x_2, \ldots, x_{|\mathbf{x}|}\}$ forward and backward. The forward GRU generates output $\mathbf{h}^{(f)}$ while the backward GRU generates output $\mathbf{h}^{(b)}$. $h_t^{(f)} \in \mathbf{h}^{(f)}$ is the output of $\mathbf{x}_{1:t}$, while $h_t^{(b)} \in \mathbf{h}^{(b)}$ is the output of $\mathbf{x}_{t:|\mathbf{x}|}$. Then $\mathbf{h}^{(f)}$ and $\mathbf{h}^{(b)}$ are concatenated to get $\mathbf{h}$ as

$$h_t = h_t^{(f)} \oplus h_t^{(b)}, \quad (6)$$

where $\oplus$ is concatenation operation of two vectors. The feature vector $v$ is the average of output $\mathbf{h}$. Formally,

$$v = \frac{1}{|\mathbf{x}|} \sum_{t=1}^{|\mathbf{x}|} h_t. \quad (7)$$

**Decoder** The decoder is a special GRU which the output is fed forward to next unit as input. Specifically, for $t$-th unit, the input is the concatenation of $v$ and $y_{t-1}$, where $y_{t-1}$ is the previous output element.

Then output element $y_{t-1}$ is selected from a dictionary according to the hidden state of the $(t-1)$-th unit $h^{(d)}_{t-1}$ using a softmax function.

**Attention** The feature vector $v$ remains identical for each unit of the decoder in a conventional seq2seq model. The attention mechanism determines different feature $v_t$ for $t$-th unit relying on the output $y_{t-1}$ and hidden state $h^{(d)}_{t-1}$ of the previous unit.

## 3 Topic Sensitive NHG

TopicNHG includes two steps: (1) assigning topics and (2) training a topic sensitive model. For short text **x**, summary **y** and topic label $l$, the model aims to make a maximum likelihood estimation on

$$p(\mathbf{y}, l|\mathbf{x}, \theta_1, \theta_2) = p(l|\mathbf{x}, \theta_1) p(\mathbf{y}|\mathbf{x}, l, \theta_2), \quad (8)$$

where $\theta_1$ is the parameter of the LDA model and $\theta_2$ is the parameter of the topic sensitive seq2seq model.

### 3.1 Topic Assignment

Latent dirichlet allocation (LDA) model (Blei et al., 2003) is effective to learn a latent topic distribution feature for a document. Given a document set $\mathbf{d} = \{d_1, d_2, \ldots, d_n\}$, LDA makes a maximum likelihood estimation on

$$p(\mathbf{d}|\alpha, \beta) = \prod_i p(d_i|\alpha, \beta), \quad (9)$$

where $\alpha, \beta$ are the Dirichlet priors on the per-document topic distribution and per-topic word distribution respectively. Then we can use

$$p(t|\alpha, \beta, d) \quad (10)$$

to infer the topic of a document.

For simplicity, we assign each input sequence with exactly one topic as

$$l = \arg_t \max p(t|\alpha, \beta, d). \quad (11)$$

### 3.2 Topic Sensitive Model

Fig. 2 shows the framework of TopicNHG. In our model, the topic label of the input short text will affect weight matrices in encoder, decoder and attention layer. For $K$ topics, we fork the model into $K$ different encoders, decoders and attention layers. With $K$ different encoder, we generate $K$ representations of input sequence $\{\mathbf{h}^{(1)}, \mathbf{h}^{(2)}, \ldots, \mathbf{h}^{(K)}\}$. Then we select a representation according to the topic label $l$ of the input sequence, where $l = \{1, 2, \ldots, K\}$. We feed $\mathbf{h}^{(l)}$ into decoders with corresponding attention layers and generate $K$ sequences of output $\{\mathbf{y}^{(1)}, \mathbf{y}^{(2)}, \ldots, \mathbf{y}^{(K)}\}$. We use $\mathbf{y}^{(l)}$ as our final output.

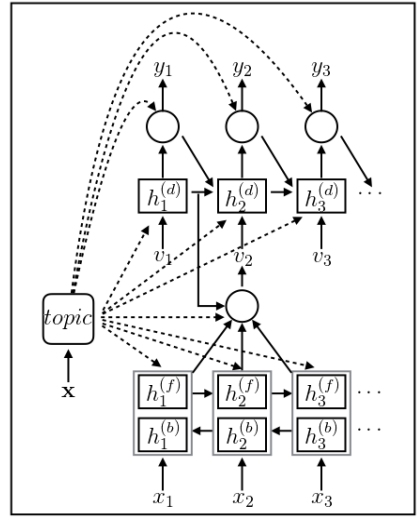

**Figure 2:** Topic Sensitive Neural Headline Generation.

The training of our model is time-consuming. For each topic, it requires 500k iterations of training. So we use the parameters trained in the conventional NHG model to initialize our model. Then we train 50k iterations for each topic to make the model topic-specific. Using initialization also makes our model more general that each part of our model will firstly be trained by a large set of general text summaries and later be trained to generate topic-specific ones.

## 4 Experiments

We conduct experiments on a large-scale Chinese short text summarization dataset (LCSTS) (Baotian Hu, 2015) to evaluate the performance of TopicNHG. LCSTS consists of short news articles with headlines from Sina Weibo*, a Chinese social media. LCSTS collects those news weibos from the verified organizations such as China Daily to guar-antee the quality of data.

---
*The website is http://weibo.com/

LCSTS contains 3 parts. Each pair in PART-II and PART-III has a human-labeled score, reflecting the relevance of the summary. The scores range from 1 to 5, 1 means 'the least relevant', while 5 means 'the most relevant'. Data in PART-II are labeled by 1 annotator, and data in PART-III are labeled by 3 annotators.

In our experiment, we use PART-I as our training data, and we use PART-III whose scores are higher than or equal to 3 as test data. Note that we take Chinese characters as input to avoid errors caused by word segmentation.

### 4.1 Results on LCSTS

We use LDA to categorize each weibo into 5 topics. Table 1 shows the most related words on each topic. We manually mark these topics as Job, Economy, Accident, Politics and Technology. Table 2 shows the number of weibos belonging to each topic.

| Topic | Keywords |
|---|---|
| Job | undergraduate, graduate, photographer, researcher |
| Economy | RMB, USD, realty, investor, company, manager, IPO |
| Accident | suspect, police, court, ID card, bus, taxi, high way |
| Politics | state department, authority, civil servants, urbanization |
| Technology | Internet, consumer, smart phone, e-business, APP |

Table 1: Topic and Keywords.

| Topic | PART-I | PART-II | PART-III |
|---|---|---|---|
| Job | 602413 | 1855 | 195 |
| Economy | 504762 | 3103 | 306 |
| Accident | 511934 | 1344 | 135 |
| Politics | 400745 | 1713 | 210 |
| Technology | 380737 | 2651 | 260 |

Table 2: Samples in Topic.

We compare our model with conventional seq2seq model(Baseline) and CopyNet (Gu et al., 2016). Table 3 shows the ROUGE-F evaluation on these models. By introducing topics, our topic sensitive model gain a 4% improvement on all 3 indexes. On each topic, we compare our model against the baseline. Table 4 shows the evaluation on each topic.

|  | Rouge-1 | Rouge-2 | Rouge-L |
|---|---|---|---|
| Baseline | 34.7 | 22.9 | 32.5 |
| CopyNet | 34.4 | 21.6 | 31.3 |
| Topic-5 | **38.4** | **26.6** | **36.1** |

Table 3: ROUGE-F(%) on LCSTS Dataset.

|  | Rouge-1 | Rouge-2 | Rouge-L |
|---|---|---|---|
| Job | 34.3/39.1 | 22.7/26.8 | 31.9/36.1 |
| Economy | 36.0/38.3 | 23.8/26.0 | 33.8/36.2 |
| Accident | 36.3/40.9 | 24.3/28.1 | 34.1/38.4 |
| Politics | 33.1/39.6 | 20.8/28.4 | 30.4/37.5 |
| Technology | 33.7/35.6 | 22.8/24.8 | 31.7/33.4 |

Table 4: ROUGE-F(%) on Each Topic (Baseline/TopicNHG).

### 4.2 Case Study

According to Table 4, we can see that short texts related to politics are most significantly improved. Fig. 3 shows an example. In this example, Topic-NHG performs much better than the baseline. The baseline considers the first sentence of the input as its main point. In most cases, this assumption is probably true. So the baseline learns a general regularity in summarization that the first sentence is important. However, the first sentence of a weibo concerning politics usually talks about a conference, while the content of the conference in the following sentences is more important. TopicNHG can figure out this regularity and give an informative summary.

---

**Weibo Text**
今天下午，北京市十四届人大三次会议召开第一场新闻发布会，聚焦老百姓"住"的问题。北京市从90年代推出、均价始终维持在每平方米4000元左右的经济适用房将成为历史。本市今年解决所有轮候家庭后，不再新建经济适用住房。
This afternoon, the 3rd meeting of Beijing's 14th Session of the National People's Congress held the first press conference, focusing on people's house problem. Beijing's economical departments with an average of ¥4000 per square meter will become history. After solving all the waiting families, no new economical departments will be built.
**Human Notation**
4000元经济适用房今年退出北京历史
¥4000 economical departments will become history this year.
**Baseline**
北京市十四届人大三次会议召开第一场新闻发布会
The 3rd meeting of Beijing's 14th Session of the National People's Congress held a press conference.
**TopicNHG**
北京经济适用房将成历史
The economical department in Beijing will become history.

Figure 3: Comparing TopicNHG with Baseline.

## 5 Conclusion and Future Work

In this paper, we propose topic-sensitive neural headline generation. The experiments prove that topic is an important feature in headline generation tasks. However, our model is relatively simple. Its high cost in training prevents it from handling more topics. In the future, we will work on the following aspects: (1) Propose a model which can easily handle plenty of topics. (2) Evaluate the performance of TopicNHG with a different number of topics. (3) Assign one short text with a distribution of multiple topics to alleviate the potential errors introduced in the topic assignment.